\pdfoutput=1
\documentclass[11pt,a4paper]{article}
\usepackage{times,latexsym}
\usepackage{url}
\usepackage[T1]{fontenc}
\usepackage{amsmath}
\usepackage{booktabs}       
\usepackage{amsfonts}       
\usepackage{nicefrac}       
\usepackage{microtype}      
\usepackage{graphicx}
\usepackage{natbib}
\usepackage{doi}
\usepackage{xcolor}
\usepackage{soul}
\usepackage{forest}
\usepackage{array}
\usepackage{paralist}
\usepackage{tabularray}
\usepackage{multirow}
\usepackage{cancel}
\usepackage{caption}
\usepackage{ulem}
\usepackage{float}
\usepackage{enumitem}
\usepackage{etaremune}
\usepackage{hyperref}
\definecolor{darkblue}{rgb}{0, 0, 0.5}
\hypersetup{colorlinks=true,citecolor=darkblue, linkcolor=darkblue, urlcolor=darkblue}

\newcommand{\llms}{LLMs\xspace}
\DeclareMathOperator*{\argmax}{arg\,max}
\DeclareMathOperator*{\argmin}{arg\,min}
\DeclareMathOperator{\EX}{\mathbb{E}}

\definecolor{midnightblue}{rgb}{0.6, 0.2, 0.3}
\definecolor{arylideyellow}{rgb}{0.6, 0.6, 0.1}

\usepackage[acceptedWithA,nohyperref]{tacl2021v1}

\usepackage{xspace,mfirstuc,tabulary}

\newif\iftaclinstructions
\taclinstructionsfalse 
\iftaclinstructions
\renewcommand{\confidential}{}
\renewcommand{\anonsubtext}{(No author info supplied here, for consistency with
TACL-submission anonymization requirements)}
\newcommand{\instr}
\fi

\iftaclpubformat 

\else

\fi

\forestset{
  default preamble={
      for tree={
            rounded corners,
            node options={align=center,},
            text width=2.7cm,
            s sep=6pt,
            calign primary child={(n_children()+1)/2},
            if n children=3 {
                calign primary child={(n_children())/2},
            }{},
            child anchor=west,
            parent anchor=east,
            grow'=east,
            draw,
            rounded corners,
            node options={
                align=center },
            text width=2.7cm,
            anchor=west,
            edge path={
                \noexpand\path[\forestoption{edge}]
                (.child anchor) -| +(-5pt,0) -- +(-5pt,0) |-
                (!u.parent anchor)\forestoption{edge label};
            },
            if level=1{
                text width=2.7cm, 
            }{},
            if level=2{
                text width=3cm, 
            }{},
            if n children=0{                
                node options={align=left, inner xsep=6pt},
                if level=2{
                    text width=8.9cm, 
                }{},
                if level=3{
                    text width=6.2cm, 
                }{},
            }{} 
    },
  }
}

\title{Bridging the Gap: A Survey on Integrating (Human) Feedback for\\ Natural Language Generation}

\author{
\textbf{Patrick Fernandes}$^{1,2,3}$  \quad
\textbf{Aman Madaan}$^{1}$ \quad
\textbf{Emmy Liu}$^{1}$  \quad
\textbf{António Farinhas}$^{2,3}$  \\ 
\textbf{Pedro Henrique Martins}$^{4}$  \quad
\textbf{Amanda Bertsch}$^{1}$ \quad
\textbf{José G. C. de Souza}$^{4}$ \quad
\textbf{Shuyan Zhou}$^{1}$ \\
\textbf{Tongshuang Wu}$^{1}$ \quad 
\textbf{Graham Neubig}$^{1, 5}$ \quad
\textbf{André F. T. Martins}$^{2,3,4}$ \quad
\\
$^1$Carnegie Mellon University\quad
$^2$Instituto Superior Técnico  (Lisbon ELLIS Unit)\quad\\
$^3$Instituto de Telecomunicações\quad
$^4$Unbabel \quad 
$^5$Inspired Cognition \quad 
\\
{\small \texttt{pfernand@cs.cmu.edu}} 
}

\begin{document}

\maketitle

\begin{abstract}
Many recent advances in natural language generation have been fueled by training large language models on internet-scale data. However, this paradigm can lead to models that generate toxic, inaccurate, and unhelpful content, and automatic evaluation metrics often fail to identify these behaviors. As models become more capable, \textit{human feedback} is an invaluable signal for evaluating and improving models. This survey aims to provide an overview of the recent research that has leveraged human feedback to improve natural language generation. First, we introduce an encompassing formalization of feedback, and identify and organize existing research into a taxonomy following this formalization. Next, we discuss how feedback can be described by its format and objective, and cover the two approaches proposed to use feedback (either for training or decoding): directly using the feedback or training \textit{feedback models}. We also discuss existing datasets for human-feedback data collection, and concerns surrounding feedback collection. Finally, we provide an overview of the nascent field of \textit{AI feedback}, which exploits large language models to make judgments based on a set of principles and minimize the need for human intervention. 
\end{abstract}

\section{Introduction}

For generation systems to be widely useful, they must generate text that is not only fluent and high-quality, but also closely aligned with human desires and specifications~\citep{vamplew2018human,hendrycks2020aligning,kenton2021alignment,turner2022formalizing,ngo2022alignment}. Achieving such ambitious goals requires modern large language models (LLMs) to evolve beyond traditional training methods. Recent improvements in this space have centered on incorporating human feedback~\citep{bai2022constitutional,ouyang2022training,openai2023gpt4}.
This feedback serves as a guiding force, steering LLMs toward the desired outcomes, much like feedback mechanisms in physical machines~\citep{aastrom2021feedback}. 

Typically, state-of-the-art language generation systems are obtained by training \textit{probabilistic}, \textit{autoregressive} LLMs on massive amounts of data using \textit{maximum likelihood estimation} (MLE). However, the data used to train these models is generally scraped from the Internet, often containing noise, social biases, and errors \citep{bolukbasi2016man,dodge-etal-2021-documenting}. This, when combined with the objective of maximizing the probability of the next token given the previous ones, might result in a \textit{misspecification}  of target behavior \citep{kenton21alignment}, and might lead to models that generate toxic, inaccurate, and unhelpful content \citep{sheng-etal-2019-woman, bender_parrots}.

\definecolor{skyblue}{RGB}{135, 206, 250}
\definecolor{bluishgreen}{RGB}{50, 205, 50}
\definecolor{yellow}{RGB}{255, 215, 0}
\definecolor{blue}{RGB}{30, 144, 255}
\definecolor{vermillion}{RGB}{255, 69, 0}
\definecolor{reddishpurple}{RGB}{221, 160, 221}
\definecolor{saffron}{RGB}{255, 165, 0}





\begin{figure*}[!htb]
\footnotesize
\begin{forest}
for tree={
    calign=center
},
[Format (\S\ref{subsec:format}), draw=skyblue, fill=skyblue!20
    [Numerical, draw=skyblue, fill=skyblue!10,
        [\citet{kreutzer-etal-2018-neural, liu-etal-2018-dialogue, fernandes-etal-2022-quality}, draw=skyblue!10, fill=skyblue!10]      
    ],
    [Ranking, draw=skyblue, fill=skyblue!10,
        [\citet{stiennon-etal-2020-learning, ouyang2022training, bai2022training}, draw=skyblue!10, fill=skyblue!10]                        
    ],
    [Natural Language, draw=skyblue, fill=skyblue!10,
        [\citet{lidialogue, madaan2023selfrefine, scheurer2023training} , draw=skyblue!10, fill=skyblue!10] 
    ],
    [Others, draw=skyblue, fill=skyblue!10,
        [\citet{mqm, pal-etal-2016-neural, nguyen-etal-2022-make}, draw=skyblue!10, fill=skyblue!10] 
    ],
]
\end{forest} \\\\
\begin{forest}
for tree={
    if level=1{ 
        text width=2cm, 
    }{},
    if n children=0{
        if level=2{
            text width=9.6cm
        }{},
        if level=3{ 
            text width=6cm, 
        }{}
    }{}
},
[Objective (\S\ref{subsec:objective}), draw=bluishgreen, fill=bluishgreen!20
    [Helpfulness, draw=bluishgreen, fill=bluishgreen!10,
        [Task Performance, draw=bluishgreen, fill=bluishgreen!10,
            [\citet{kreutzer-etal-2018-neural, stiennon-etal-2020-learning}, draw=bluishgreen!10, fill=bluishgreen!10]
        ],
        [Instruction-Following, draw=bluishgreen, fill=bluishgreen!10,
            [\citet{ouyang2022training, askell2021general}, draw=bluishgreen!10, fill=bluishgreen!10]         
        ]             
    ],
    [Harmlessness, draw=bluishgreen, fill=bluishgreen!10,
        [\citet{ouyang2022training, bai2022training, bai2022constitutional, glaese2022improving}, draw=bluishgreen!10, fill=bluishgreen!10]                        
    ]
]
\end{forest}\\\\
\begin{forest}
for tree={
    if level = 1 {
        text width=2.5cm
    }{},
    if n children=0{ 
        text width=5.5cm, 
    }{}
},
[Usage, draw=vermillion, fill=vermillion!20 
    [Training {(\S\ref{subsec:direct-optimize}}{,\S\ref{subsubsec:indirect-optimize})}, draw=vermillion, fill=vermillion!10
        [Feedback-Based Imitation Learning, draw=vermillion, fill=vermillion!10
            [\citet{lidialogue, glaese2022improving, scheurer2023training}, draw=vermillion!10, fill=vermillion!10, ]
        ],
        [Joint Feedback Modelling, draw=vermillion, fill=vermillion!10,
            [\citet{lidialogue, hancock2019learning, korbak2023pretraining, yuan2023rrhf}, draw=vermillion!10, fill=vermillion!10, ]
        ],
        [Reinforcement Learning, draw=vermillion, fill=vermillion!10,
            [\citet{kreutzer-etal-2018-neural, stiennon-etal-2020-learning, askell2021general}, draw=vermillion!10, fill=vermillion!10, ]
        ],
    ],
    [Decoding {(\S\ref{subsec:direct-decode}}{,\S\ref{subsubsec:indirect-decode})}, draw=vermillion, fill=vermillion!10
        [Reranking, draw=vermillion, fill=vermillion!10
            [\citet{fernandes-etal-2022-quality, gao2022scaling}, draw=vermillion!10, fill=vermillion!10]
        ],
        [Feedback-Conditioning, draw=vermillion, fill=vermillion!10
            [\citet{Schick2022PEERAC, madaan-etal-2022-memory}, draw=vermillion!10, fill=vermillion!10]
        ]
    ]
]
\end{forest} \\\\
\begin{forest}
for tree={
            if level=1{
                text width=3.6cm, 
            }{},
            if level=2{
                text width=3cm, 
            }{},
            if n children=0{ 
                if level=2{
                    text width=8.1cm, 
                }{},
                if level=3{
                    text width=6.5cm, 
                }{},
            }{}
    },
[Modeling, draw=saffron, fill=saffron!20
    [\textit{None} {(\S\ref{sec:direct-opt})}, draw=saffron, fill=saffron!10
        [\citet{lidialogue, kreutzer-etal-2018-neural, madaan-etal-2022-memory}, draw=saffron!0, fill=saffron!10]                      
    ],
    [Feedback-Modeling {(\S\ref{sec:indirect})}, draw=saffron, fill=saffron!10
        [\citet{gao-etal-2018-april, stiennon-etal-2020-learning, bai2022training}, draw=saffron!0, fill=saffron!10]
    ],
    [AI Feedback {(\S\ref{sec:ai-feedback})}, draw=saffron, fill=saffron!10
        [\citet{Yang2022Re3GL, bai2022constitutional, madaan2023selfrefine}, draw=saffron!0, fill=saffron!10]                         
    ]
]
\end{forest}
\caption{Taxonomy of methods that leverage human-feedback, with some example representative works in the literature that fit in each category.}
\vspace{-1.1em}
\label{fig:taxo-human-feedback}
\end{figure*}
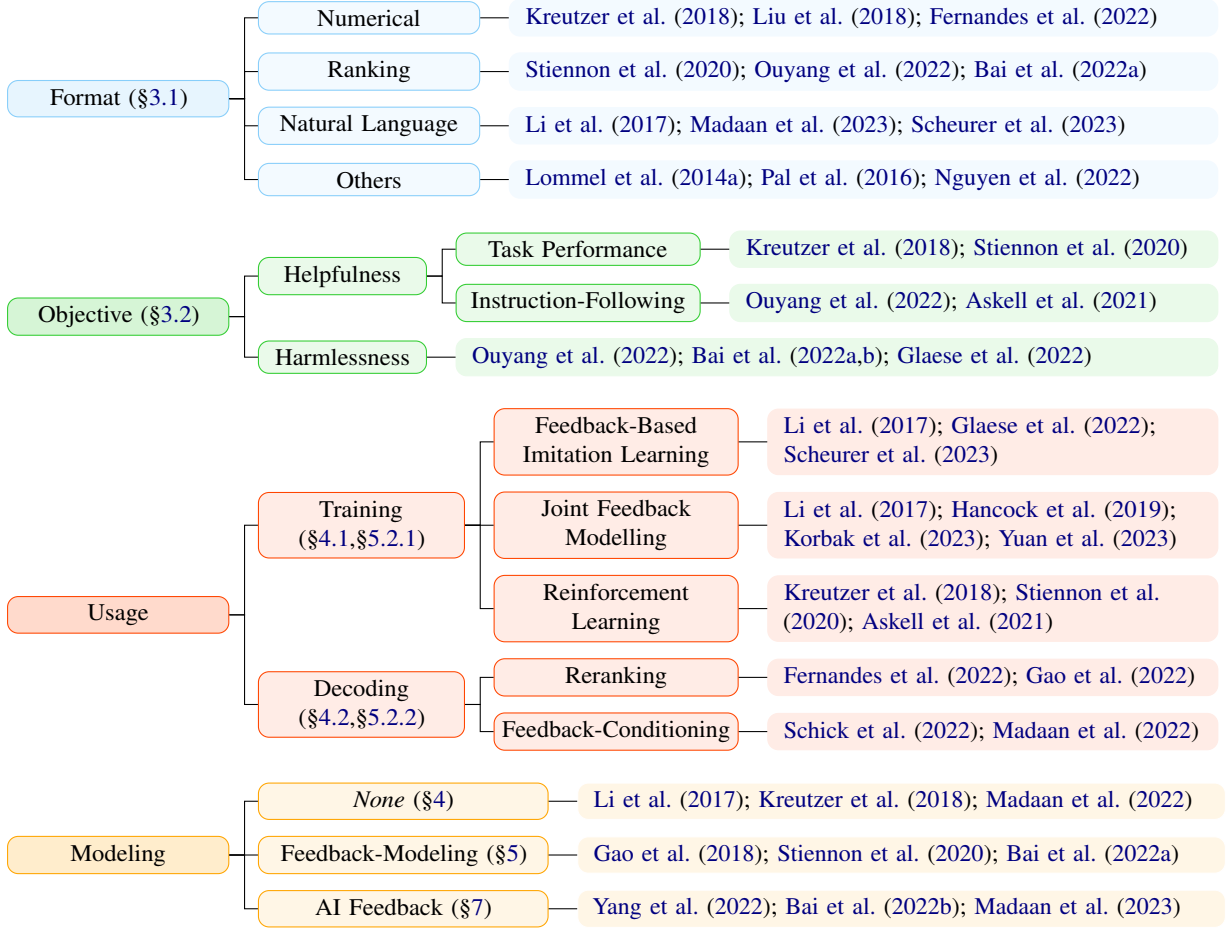

Exacerbating the problem above is the fact that these models are often evaluated using automatic metrics that compare the generated text with some ``reference'' text using surface-level features (such as word overlap), which often do not correlate with \textit{human-perceived} quality of text \citep{ schluter-2017-limits, mathur-etal-2020-tangled, Gehrmann2022GEMv2MN}, especially when models are optimized for them \citep{Dpaulus17summarization, amrhein-sennrich-2022-identifying}. This difficulty in evaluation arises partly because, for many tasks, there is not a single correct answer since the same communicative intent can be conveyed in multiple ways. 

Leveraging human assessments to evaluate the quality of texts generated by models is then a popular approach. Crucially, considering human-perceived quality can help close the \textit{gap} between machine and human generated text, and help in addressing the challenges posed by \textit{Goodhart's law}: ``when a measure becomes a target, it ceases to be a good measure'' \citep{Goodhart1984}. 
This realization has spurred a growing interest in improving natural language generation systems by leveraging \textit{human feedback} on model-generated outputs, and has led to the emergence of the first widely-used general-purpose language assistants \citep{openai2023gpt4}.
Human feedback not only enhances system performance, but also serves as a mechanism to steer the system in alignment with desired outcomes or goals~\citep{rosenblueth1943behavior,wiener1948cybernetics}. \looseness=-1

Feedback, as a concept, encompasses a wide range of meanings and interpretations~\citep{wiener1948cybernetics}; however, some universal characteristics can be identified, such as its format, its intended results, and the ways it is utilized as a part of the model development process. In this survey, we focus on the role of \textit{human feedback} for improving language generation.
We start by formalizing the notion of \textit{human feedback} and creating a taxonomy of the different types of feedback in the literature, and of how they have been used (\S\ref{sec:hf-types}). We discuss how we can describe feedback by its \textit{format} and its \textit{objective}, in terms of the desired model behavior (\S\ref{sec:describing-feedback}). We discuss approaches that directly optimize models against human feedback on (their) outputs, for example, using reinforcement learning with human reward functions (\S\ref{sec:direct-opt}). We then move to approaches that circumvent the costs of direct feedback optimization by first training \textit{feedback models} to approximate human feedback, and then improving generation using these proxy models (\S\ref{sec:indirect}). We discuss existing datasets for human-feedback data, how these datasets are typically collected, and the impact that the collection process might have on the behaviour of the models (\S\ref{sec:collection}). Finally, we discuss a recent line of work that reduces the need to collect human feedback by leveraging \textit{AI feedback} from large language models (\S\ref{sec:ai-feedback}).

\section{A Taxonomy for Leveraging (Human) Feedback for Generation}
\label{sec:hf-types}

\subsection{Background}
\label{sec:background}

Consider a model $M: \mathcal{X} \rightarrow \mathcal{Y}$ which, given an input of some type $x \in \mathcal{X}$, outputs text $\hat{y} \in \mathcal{Y}$. Importantly, while $x$ can be of any format, we restrict ourselves to cases where $y$ is in the space of \textit{natural language} (\textit{i.e.}, $\mathcal{Y} \subseteq \Sigma^\star$ for some alphabet $\Sigma$).
This general formulation encompasses a wide range of NLG tasks. For example:

\begin{itemize}
    \item \textbf{Summarization}: $\mathcal{X}$ is the space of documents, and $\mathcal{Y}$ the space of possible summaries.
    \item \textbf{Machine Translation}: $\mathcal{X}$ and $\mathcal{Y}$ are the spaces of sentences in the source and target languages, respectively.
    \item \textbf{Dialog Generation}: $\mathcal{X}$ is the space of possible dialog histories, and $\mathcal{Y}$ is the space of possible responses.
    \item \textbf{Image Captioning}: $\mathcal{X}$ is the space of images, and $\mathcal{Y}$ is the space of possible captions.
\end{itemize}
These models are generally realized as a parameterized, conditional probability distribution $P_{\theta}(y|x)$, where $\theta$ are the model parameters. This distribution is often estimated autoregressively: the probability of a sentence $y$ given an input $x$ is decomposed into the product of the probabilities of each token in the sentence, conditioned on the previous tokens. 
These models are then trained by finding the parameters $\theta^\star$ that maximize the likelihood of some training data $\mathcal{D} = \{(x_i, y_i)\}_{i=1}^N$.
Then, at \textit{inference} time, given an input $x$, an output $\hat{y}$ is decoded from the learned distribution. 
This decoding can be done, for example, by approximating the most-likely sequence of tokens ($M(x) \approx \argmax_{y} P_{\theta^\star}(y|x)$) or by random sampling ($M(x) \sim P_{\theta^\star}(y|x)$).

Evaluating the quality of generated text $\hat{y} \in \mathcal{Y}$ can be challenging due to the complexity and subjectivity of natural language. Various automatic metrics have been proposed for various domains/tasks. These metrics traditionally rely on n-gram matching or other simple heuristics that cannot account for complex linguistic phenomena (such as paraphrasing or stylistic variations) and often fail to capture all the nuances of human judgment \citep{10.1145/3485766,Gehrmann2022GEMv2MN}. For this reason, for many of these tasks, asking for \textit{human feedback} is considered the gold standard for assessing the quality of the generated text, and newer \textit{learned} metrics often aim to approximate the way humans provide feedback (see \S\ref{subsec:preference-models}).

More formally, we consider \textbf{human feedback} to be a family of functions $\mathcal{H}$ such that each \textit{feedback function} $h \in \mathcal{H}$ takes an input\footnote{Although feedback can be provided independently of the input (for example for \textit{fluency}), we assume some (potentially empty) input for simplicity of notation.} $x\in \mathcal{X}$ and one or more outputs $y_1, \cdots, y_n \in \mathcal{Y}$ and returns some \textit{feedback} $f \in \mathcal{F}$:
\begin{equation}
h : \mathcal{X} \times \underbrace{\mathcal{Y}_1 \times \cdots \times \mathcal{Y}_n }_n \rightarrow \mathcal{F}.
\end{equation}
A simple example of a (human) feedback function is asking humans to say if, given an input, a particular output is good or bad ($h: \mathcal{X} \times \mathcal{Y} \rightarrow \{0,1\}$). However, more complex feedback functions, such as rankings or natural language feedback, exist and are commonly used (see \S\ref{subsec:format}). 

We note that this framing is a \textit{simplification} of the real world: often, different humans might provide different (and potentially contradicting) feedback for the same outputs, and a single function might not be able to capture this variability in human opinion (we discuss this further in \S\ref{sec:collection}). Finally, while our formalization is flexible, it excludes other approaches where models interact with humans to improve learning, such as active learning and other \textit{human-in-the-loop} approaches.

\subsection{Taxonomy}

Having established a basic mathematical formulation, we now identify four key axes along which we can classify the uses of human feedback:

\paragraph{What is the \textit{format} of the feedback?} The format of human feedback can vary, including binary judgments, numerical scores, ordinal rankings, or qualitative natural language explanations.

\paragraph{What is its \textit{objective}?} Depending on the use case of our model, the feedback can have a variety of purposes, ranging from assessing model performance and accuracy to preventing toxicity and harmful behavior.

\paragraph{When is it \textit{used}?} Human feedback can be incorporated into the training stage to optimize the model parameters directly. Alternatively, it can be used at inference time to guide the decoding process.

\paragraph{How is it \textit{modeled}?} While ideally, we would use direct feedback from humans whenever possible, the prohibitive cost of its collection means that it is often useful to instead use \textit{surrogate} models that approximate human preferences.

\begin{table*}[h]
\centering
\small
\begin{tblr}{
  colspec={X[4, c]X[4,c]X[4,c,m ]X[2,c,m]},
  rowsep=0.2em,
  colsep=0.2mm
}
\hline
Input & Output(s) & Feedback & Type \\ \hline
\SetCell[r=3]{c} {\textit{A melhor comida do mundo é} \\ \textit{a portuguesa.}} & \SetCell[r=3]{c} {\textit{The worst food in the world} \\ \textit{are Portuguese.}} & 0.7 & Score \\
\cline{3, 4}
& & {'worst': \texttt{\small major/accuracy}  \\ 'are': \texttt{\small minor/fluency}} & MQM \\
\cline{3, 4}
& & 'worst' $\rightarrow$ 'best', 'are' $\rightarrow$ 'is' & Post-Edition \\ \hline
\SetCell[r=2]{c} {\textit{Artificial intelligence has the} \\ \textit{potential to revolutionize} \\\textit{industries (...) but ethical} \\ \textit{concerns need to be handled.}} & \SetCell[r=2]{c} \textit{AI can change industries.} & {\texttt{Fluency}: 1 \\ \texttt{Relevance}: 0.7}  & { Multi-Aspect} \\
\cline{3, 4}
& & \textit{"Misses the ethical concerns."} & Natural Language \\ \hline
\SetCell[r=2]{c} {\textit{Explain the moon landing} \\ \textit{to a 6 year old}} & A: \textit{People went to the ...} & \SetCell[r=2]{c} A > B & \SetCell[r=2]{c} Ranking \\
\cline{2-2}
& B: \textit{The moon is a satellite...} & & \\
\hline
\end{tblr}
\caption{\label{table:feedback} Example input and output for three tasks (machine translation, summarization, and instruction following) and possible different (example) feedback that can be given.}
\vspace{-1em}
\end{table*}

\section{Describing Feedback}
\label{sec:describing-feedback}
\subsection{Format}
\label{subsec:format}

An important decision to make when we want to improve language generation systems through human feedback is in what \textit{format} to collect this feedback in. The choice of format has implications on the expressivity of the feedback, the ease of its collection, and how we can use it to improve systems.
In particular, the complexity of the feedback format is an important factor: simpler formats are often easier to collect and use as part of the training/decoding process, but contain less information than more ``complex'' formats, and might not be able to capture important information for improving the system. The choice of format also has implications in the difficulty for humans to give feedback, its consistency/agreement, and the level of \textit{rationality} of said feedback \citep{ghosal2023effect}. Types of feedback are summarized in \autoref{table:feedback} with examples. \looseness=-1

\paragraph{Numerical}
Numerical feedback, which takes an input and output and returns a single score ($\mathcal{X} \times \mathcal{Y} \rightarrow \mathcal{N} \subseteq \mathbb{R}$), is one of the simplest feedback formats to collect and use.
\citet{kreutzer-etal-2018-neural} studied using \textit{categorical} feedback, in the form of 5 possible ``stars'' that can be assigned to a translation, which are then averaged to produce a score ($\mathcal{N} = [1, 5]$) and used to improve the model. \citet{liu-etal-2018-dialogue} and \citet{shi-etal-2021-refine-imitate} used even simpler feedback, by asking humans to choose if a given response is good or not ($\mathcal{N} = \{0,1\}$). Numerical feedback has also been extensively used for evaluation, albeit not with the explicit goal of improving generation. For example, \textit{direct assessments} \citep{graham-etal-2013-continuous} in machine translation ask humans to rate translations on a continuous scale, and some works have attempted to use this feedback data to train feedback models \citep{sellam-etal-2020-bleurt, rei-etal-2020-comet} and improve generation \citep{freitag-etal-2022-high, fernandes-etal-2022-quality}. 

Although easy to leverage, numerical feedback suffers from some limitations: depending on the complexity of the generation task, reducing feedback to a single score might generally be a hard and ill-defined task for humans, leading to a costly collection process and problems of \textit{subjectivity} and \textit{variance} (see \S\ref{subsubsec:subjectivity-variance}). Furthermore, such feedback might not be suited to distinguish between outputs of similar quality.

\paragraph{Ranking-based}
An alternative to asking humans to assign a single score to a given input-output pair is asking them to \textit{rank} multiple possible alternative outputs 
$$h : \mathcal{X} \times \mathcal{Y}_1 \times \cdots \times \mathcal{Y}_n  \rightarrow S_n$$
where $S_n$ represents the set of all permutations/rankings of $n$ elements (optionally allowing ties). This has been used extensively in evaluation  \citep{chaganty-etal-2018-price}. Compared to numerical feedback, this format tends to be easier to collect, and, potentially, for this reason, ranking-based feedback tends to be collected to improve model behavior rather than just for evaluation (since the former tends to require more feedback data). \citet{ziegler-etal-2019-fine} and \citet{stiennon-etal-2020-learning} asked humans to rank alternative summaries of the system they are trying to improve. 
Similarly, \citet{ouyang2022training} collected rankings of alternative responses to an \textit{instruction} given to the model. They utilized these rankings to enhance the model's \textit{instruction-following} capabilities. Subsequent research has also employed ranking-based feedback for the same task \citep{askell2021general, bai2022training, bai2022constitutional}.

\paragraph{Natural Language}
Both numerical and ranking-based feedback lack the ability to capture detailed information about problems with the output, which can be crucial for improving generation systems. Instead of asking humans to rank or score outputs, we can instead ask for \textit{natural language} feedback. In such cases, the feedback typically provides more detailed information, either highlighting the shortcomings of the current output or suggesting specific actions for improvement. For example, \citet{lidialogue} asked humans to give natural language feedback to a dialogue question answering model, including positive or negative feedback, but also possibly providing the correct answer to the model or hinting about it.
\citet{tandon-etal-2022-learning} and \citet{madaan-etal-2022-memory} gather natural language feedback on errors present in model-generated graphs and the model's interpretation of a given instruction.
\citet{scheurer2022training, scheurer2023training} improve summarization capabilities of language models by asking humans to provide natural language feedback of summaries of the model. \citet{Li_2022} collect natural language feedback (in addition to numerical feedback) for responses from a Question Answering (QA) system.

\paragraph{Others}
Besides these feedback types, other (potentially domain-specific) types of feedback can be used to improve model behavior. Commonly humans are asked to provide \textit{multi-aspect} feedback ($\mathcal{X} \times \mathcal{Y} \rightarrow \mathbb{R}^d$ or $\mathcal{F}^d$ more generally), scoring an output or ranking multiple outputs with respect to multiple dimensions \cite{bohm-etal-2019-better,glaese2022improving,madaan2023selfrefine, nguyen-etal-2022-make}. \textit{Post-editions} ask humans to provide corrections to the output in the form of small edits (\textit{e.g.}, \textit{replace X by Y}), and post-edition data has been used to directly improve models \citep{denkowski-etal-2014-learning} or train \textit{automatic post edition} systems that correct model mistakes \citep{pal-etal-2016-neural,mehta2019improving,madaan2021graphcorr,talmor2020teaching,elgohary2021nl}. There are also other feedback types that haven't been fully leveraged to improve generation: \textit{e.g.}, \textit{Multidimensional Quality Metrics (MQM)} \citep{lommel2014multidimensional}, the standard for evaluating translation quality, asks professional translators to identify errors \textit{spans} in a translation, alongside severity and type of error.


\subsection{Objective}
\label{subsec:objective}

The purpose of collecting feedback is to \textit{align} the model's behavior with some (often ill-defined) \textit{goal} behavior: we might want our summarization model to generate summaries that contain all core information, even if it means they are a bit longer; in \textit{commercial} machine translation, extra care is given to ensure that models do not mistranslate business-critical information; and in dialogue agents, we might want the model to be able to produce polite and harmless responses. This \textbf{alignment objective} has been studied extensively in the \textit{AI safety and alignment} literature \cite{bostrom_alignment, amodei2016alignment, Bommasani2021OnTO}. In addition, \citet{kenton21alignment} discuss some behavioral issues in language agents (natural language generation models) arising from a \textit{misspecified} alignment objective (for example, from noisy labels in the training data), and \citet{Leike2018ScalableAA} proposed using feedback models to tackle the difficulty in specifying this objective.

\citet{bai2022training} explicitly divided the problem of ``aligning'' a language model into improving its \textbf{helpfulness} and increasing its \textbf{harmlessness}. Most works implicitly consider either the use of feedback that targets performance factors (such as when targeting overall performance in a task or ability to follow instructions) or harmlessness factors (such as not producing toxic text or providing information that could lead to harm).\footnote{We mostly ignore the proposed \textit{honesty} aspect, as none of these works tackle this directly.}

\paragraph{Helpfulness}
Most often, feedback is collected with some \textit{helpfulness} objective in mind: a necessary (but not sufficient) condition for a helpful system is that it performs the task well, and so feedback related to \textbf{task performance} generally falls under this umbrella. For example, most works in machine translation leverage feedback related to the quality of translation \citep{kreutzer-etal-2018-neural, fernandes-etal-2022-quality}, which is expected to be correlated with its helpfulness in downstream applications. Similarly, in summarization, most works leverage feedback related to aspects such as \textit{relevance}, \textit{consistency} and \textit{accuracy} \citep{ziegler-etal-2019-fine, stiennon-etal-2020-learning} (in short, the quality of the summary). One particularly well-studied feedback objective is the ability to \textbf{follow instructions} \citep{ouyang2022training}: the task of instruction-following can encompass a wide range of other tasks, and using feedback to improve (instruction following) language assistants has been considered a benchmark for the alignment problem \citep{askell2021general}.

\paragraph{Harmlessness}
Another important alignment objective is \textit{harmlessness}: we want our models not to produce certain types of output or violate certain norms. Feedback collected in \citet{ouyang2022training} considered aspects such as the toxicity of text (besides the overall ability to follow instructions). \citet{bai2022training} explored the interaction between the helpfulness and harmlessness objectives, showing a trade-off between both. 
\citet{lambda2022} collected feedback on whether their model violates a set of safety objectives and used it to finetune the model.
\citet{glaese2022improving} also ask humans to provide feedback on the harmlessness of their system, by defining a set of \textit{rules} and asking humans if the outputs violate these rules.
\citet{bai2022constitutional} showed that feedback produced by LLMs could increase harmlessness without reducing helpfulness.

\section{Directly Leveraging Human Feedback}

In an ideal scenario, we would directly leverage human feedback to improve generation: humans would provide the feedback for training or decoding procedures.  

\label{sec:direct-opt}

\subsection{Optimizing for Human Feedback}
\label{subsec:direct-optimize}

Once human feedback has been collected, one way to use it is by optimizing the model parameters directly. 
However, this requires the feedback to be ``optimizable'', \textit{i.e.}, possibly formulated as an optimization problem based on which we can obtain an improved model. For instance, if the feedback is a numerical score ($f \in \mathbb{R}$), we can create the following optimization problem:
\begin{equation}
\label{eq:optimize-feedback}
\theta^\star = \argmax_\theta \EX_{x \sim \mathcal{D}} [ h(x, M_\theta(x)) ].
\end{equation}
Where $\mathcal{D}$ is the distribution of possible inputs. Various techniques have been suggested to optimize the model parameters, $\theta$, using the collected human feedback. These can be divided into three main categories based on the training mechanisms, which we will call \textbf{feedback-based imitation learning}, \textbf{joint-feedback modeling}, and \textbf{reinforcement learning (RL)}.

The \textbf{feedback-based imitation learning} approach involves using human feedback to optimize the model by performing supervised learning with a \textit{dataset} composed of positively-labeled generations together with the corresponding inputs, $\mathcal{D^+}$. This can be achieved by minimizing the loss: 
\begin{align}
\theta^\star &= \argmin_\theta \sum_{i=1}^{|\mathcal{D^+}|} \mathcal{L}^{(i)} (\theta) \\
\mathcal{L}^{(i)} (\theta) &= - \log p_\theta \left(y^{(i)} \mid x^{(i)} \right)
\end{align}
\label{eq:feedback-based-imitation}
An instance of this approach can be found in \citet{lidialogue}, in which the authors train a dialogue model by maximizing the likelihood of the model's answers labeled as correct by humans. Similarly, \citet{kreutzer-etal-2018-neural} trained a machine translation model on a set of positively-labeled translations, and \citet{glaese2022improving} performed supervised learning on the preferred dialogues which comply with their pre-defined rules (concerning correctness, harmfulness, and helpfulness), according to humans.
A slightly different approach was proposed by \citet{hancock2019learning}: deploying a chit-chat dialogue model and using the human utterances as targets to fine-tune the model.
\citet{scheurer2022training, scheurer2023training} leverage the fact that \llms can follow instructions and start by collecting natural language human feedback about the model generations, which often describes what an improved text would look like. Then, they ask the LM to generate multiple refinements based on the input, previous model generation, and the corresponding feedback. The highest similarity refinements for each generation are then used to fine-tune the LLM. 
OpenAI's \texttt{text-davinci-002} was trained with both human demonstrations and model outputs with the highest possible rating, an approach deemed \textit{FeedME} \citep{openai_model_index}.
A downside of these approaches is that they disregard the generations which do not receive positive feedback, which may contain useful information to optimize the model.

On the other hand, \textbf{joint-feedback modeling} leverages all the information collected by directly using human feedback to optimize the model. Also, as the feedback is modeled directly by the model, this approach allows feedback in formats other than numerical or ranking-based (\textit{e.g.}, natural language). 
Having $\mathcal{D}$ as the \textit{dataset} of inputs $x$, generations $y$, and human feedback $f$ collected, this can be achieved by minimizing the following loss of the form
\begin{align}
\label{eq:joint-feedback-modeling}
    \mathcal{L}^{(i)} (\theta) = - \log p_\theta \left( y^{(i)}, f^{(i)} \mid  x^{(i)} \right)
\end{align}
Over all examples in $\mathcal{D}$. These equation can be factorized as $ \mathcal{L}^{(i)} (\theta) = - \log p_\theta \left(f^{(i)} \mid y^{(i)}, x^{(i)} \right) + \log p_\theta \left(y^{(i)} \mid  x^{(i)}  \right)$. Some works simply train the model to predict the feedback given to each generation \citep[forward prediction]{weston2016dialog}, disregarding the second term of the factorization. 
One example of this approach is the work of \citet{lidialogue}, in which the authors asked humans to give natural language feedback (\textit{e.g.}, positive/negative feedback, providing the correct answer to the model, or giving a hint about the correct answer) to a dialogue question answering model. 
Then, after having collected the feedback, the model is trained to predict it.
\citet{hancock2019learning} proposed having an auxiliary model predicting the satisfaction of the human speaking with the model. Then, if the satisfaction score is lower than a pre-defined threshold, the model will ask the human for feedback. The model then leverages the natural language feedback humans give by learning to predict it.
\citet{yuan2023rrhf, rafailov2023direct} showed that having summarization models predict the rankings of different summaries helps the model generate better summaries, and might even outperform more complicated approaches using feedback models (\S\ref{sec:indirect}).

Other works train the model to predict the generations and the corresponding human feedback.
\citet{xu2022learning} proposed using the \textsc{Director} model introduced by \citet{arora2022director} to leverage human feedback. As this model has a unified decoder-classifier architecture, \citet{xu2022learning} proposed using positively-labeled examples to train its language modeling head (similarly to feedback-based imitation learning) and using both the positive and negatively-labeled examples to train a classifier head that directs the model away from generating undesirable sequences.
\citet{thoppilan2022lamda} follow this approach to enforce the model's quality and safety. First, they collect dialogues between crowd-workers and the proposed language model LaMDA, which are annotated with feedback provided by the crowd-workers. This feedback states each response's quality (sensible, specific, and interesting) or safety. Then, LaMDA is fine-tuned to predict the high-quality responses and the rewards given to every response regarding its quality attributes and safety. At inference time, LaMDA is also used to filter out candidate responses for which its safety prediction is below a threshold.

Finally, this can also be achieved by training the model to predict generation and conditioning on the feedback. This corresponds to minimizing the following loss:
\begin{equation}
\label{eq:conditional-feedback-modeling}
    \mathcal{L}^{(i)} (\theta) = -\log p_\theta \left(y^i \mid  f^i, x^i \right)
\end{equation}

\citet{liu2023languages} proposed prompt-based fine-tuning, where they create prompts containing previous generations rated by humans, in the order of preference. They also suggest inserting language-based feedback (\textit{e.g.}, ``... is a worse answer than ...'') to the prompt, between the generations. Then, the model is fine-tuned to maximize the likelihood of generating the most preferred answer.

Finally, \textbf{reinforcement learning (RL)} offers a more versatile approach, allowing for direct optimization of a model's parameters based on human feedback, regardless of the feedback's differentiability. A common RL algorithm used in this context is the REINFORCE algorithm~\citep{10.1007/BF00992696}, which updates the policy parameters using the following gradient:
\begin{equation}
\label{eq:policy-gradient}
\nabla_\theta J(\theta) = \EX_{x\sim \mathcal{D}, y\sim p_\theta}[ h(x, y) \nabla_\theta \log p_\theta(y \mid x) ]
\end{equation}
Here, $\mathcal{D}$ represents the set of inputs $x$, and $p_\theta$ is the policy. This flexibility enables RL to handle various types of feedback and better align the generated output with human preferences.
For instance, \citet{kreutzer-etal-2018-neural} proposed using task-based implicit feedback from user queries as a reward signal to train a machine translation model using a word-level variant of minimum risk training \citep{shen-etal-2016-minimum}, while \citet{jaques2019feedback} used implicit human reactions in chat to improve open-domain dialog systems through off-policy Q-learning~\citep{watkins1992q}.
Given that collecting human feedback can be expensive and time-consuming, learning is done offline from logged data, which is typically more favorable than on-policy settings that need feedback on the fly.
Later in \S\ref{subsubsec:indirect-optimize}, we discuss several works that attempt to optimize feedback models using RL instead of directly optimizing human feedback. In conjuction, these aproaches are commonly known as \textit{Reinforcement Learning from Human Feedback} (\textbf{RLHF}). \looseness=-1

\subsection{Decoding with Human Feedback}
\label{subsec:direct-decode}

While directly optimizing model parameters provides greater control, modifying them may not always be feasible, particularly in the case of \llms.
Additionally, feedback might be unavailable during model training, limiting the scope for parameter adjustments.
In such cases, leveraging human feedback during decoding plays a critical role in enhancing \llms's performance.
This type of feedback, derived from interactions between \llms and users in practical scenarios, enables models to learn from their errors and offers opportunities for ongoing refinement without altering model parameters. In addition, the feedback functions as a guiding mechanism, allowing the model to generate more desirable outputs by leveraging its existing capabilities. \looseness=-1

There are two broad categories in which human feedback is used in this setup: \begin{inparaenum} \item \textit{Feedback Memory:} Feedback Memory Utilization involves maintaining a repository of feedback from prior sessions. Then, when processing new inputs, the system uses relevant feedback from similar inputs in its memory to guide the model toward generating more desirable outputs based on past experiences and user preferences.
While a classical concept~\citep{riesbeck1981failure,schank1983dynamic}, recent work has shown the promise of such a memory-augmented approach in both  finetuning~\citep{weston2014memory,wu2018query,tandon-etal-2022-learning}  and few-shot setups~\citep{madaan-etal-2022-memory}.

\item \textit{Iterative Output Refinement:} This method employs human feedback to refine the model's output iteratively. Users can provide feedback on intermediate responses, enabling the model to adjust its output until it meets the user's satisfaction. This process allows the model to better understand user preferences and produce more suitable outcomes~\cite {reid2022learning,Saunders2022SelfCritique, Schick2022PEERAC,nijkamp2022conversational}. Feedback can also be provided on model attributes such as the decoding strategy \citep{passali-etal-2021-towards}, rather than directly on its outputs.
\end{inparaenum}

These two techniques are not mutually exclusive and can be combined to achieve even better performance, creating a more adaptive and responsive system that caters to user expectations.

\section{Improving Generation using Human Feedback Models}
\label{sec:indirect}

Directly using human feedback to improve model behavior is not feasible in the general case: asking humans to provide feedback for \textit{every} model output is both expensive and time-consuming. 

\subsection{Learning Models of Human Feedback}
\label{subsec:preference-models}

An alternative approach to obtaining human feedback is to develop models that can predict or approximate it. Although these models may not be perfect, they offer the advantage of providing feedback at a low cost after training, thereby enabling the scaling of feedback-dependent techniques.

More formally, given a feedback function $h : \mathcal{X} \times \mathcal{Y}_1 \times \cdots \times \mathcal{Y}_n  \rightarrow \mathcal{F}$, we want to learn a \textit{parametric} (numerical) feedback model $\hat{h}_\phi: \mathcal{X} \times \mathcal{Y} \rightarrow \mathbb{R}$ (with parameters $\phi$) that ``agrees'' with human feedback. 
This agreement is expressed through a loss function, and the model is trained to minimize this agreement loss: 
\begin{align}
\phi_\star &= \argmin_\phi \EX_{x, y_1, \cdots, y_n \sim \mathcal{D}_f} \left [\mathcal{L} (\phi)\right ] \\ 
\mathcal{L} (\phi) &= \mathrm{loss} \left (\hat{h}_\phi(x, y_1), \cdots, h(x, y_{1:n})\right )
\end{align}
For example, if the feedback function we are trying to model is also numerical ($h: \mathcal{X} \times \mathcal{Y} \rightarrow \mathbb{R}$), then this loss can just be any standard regression loss, such as the squared difference between the human feedback and model feedback $\mathcal{L} (\phi) = \left (\hat{h}_\phi (x, y) - h(x, y)\right)^2$. Importantly, while the feedback model is (generally) numerical, the human feedback can be in any other format, as long as a suitable loss function can be specified. \citet{stiennon-etal-2020-learning} train \textit{preference} models \footnote{We specify the feedback model with respect to the human feedback format, \textit{i.e.},  \textit{reward} and \textit{preference} model for numerical and ranking-based human feedback,  respectively.} $\hat{h}_\phi(x, y_n)$ on ranking-based feedback, using a loss of the form
\begin{equation}
\mathcal{L} (\phi) = \log \left ( \sigma \left (\hat{h}_\phi (x, y_{+1}) - \hat{h}_\phi (x, y_{-1}) \right ) \right )
\end{equation}
such that sample $y_{+1}$ was preferred to $y_{-1}$ for the same input $x$: $h(x, y_{-1}, y_{+1}) = \left ( y_{-1} < y_{+1} \right )$. Variants of this loss have subsequently been used in other works \citep{ouyang2022training, askell2021general, liu-etal-2022-brio, qin-etal-2022-t5score, yuan2023rrhf}.

The problem of feedback modeling has been studied extensively in the context of \textit{metric learning} for NLP. 
\citet{zhang2019bertscore} and \citet{zhou2023codebertscore} utilized pre-trained masked LMs to compute similarity scores between the generated text or code snippets and their references.
In MT, \citet{sellam-etal-2020-bleurt} and \citet{rei-etal-2020-comet} trained BLEURT and COMET, respectively, to regress on human quality assessments 
of translation quality.
For summarization, \citet{zopf-2018-estimating} leveraged annotated pairwise preferences 
to train a preference model and
\citet{peyrard-etal-2017-learning} learned a summary-level metric from a set of human judgements included in older summarization datasets~(\textit{e.g.,} TAC-2008).
These metrics have been shown to correlate much better with human judgments than widely used lexical-metrics such as BLEU and ROUGE \citep{freitag-etal-2022-results}.
It is notable that these reward models were not trained with the intent of improving generation directly, although some of them were used for that purpose later, as discussed in \S\ref{subsec:Leveraging Preference Models to Improve Generation}.

Recently, there has been a growing interest in developing feedback models directly with the aim of using them to improve generation \citep{bohm-etal-2019-better, ziegler-etal-2019-fine}.
As a first step, these models are typically initialized with weights from either the target LM that requires improvement or from a model of the same family~(\textit{e.g.,} of a smaller size)~\citep{askell2021general,bai2022training,ouyang2022training}. 
One key consideration in the initialization is the size of the pretrained model: while scaling up may improve overall performance~\citep{askell2021general,bai2022training}, \citet{ouyang2022training} find that larger models may be less stable for future finetuning.

Next, the feedback model is finetuned on a dataset of human feedback. This dataset is typically collected by asking annotators to provide feedback on outputs from an earlier version of the model being improved. However, it is also possible to first finetune the feedback model on naturally occurring implicit feedback, such as from user interactions on websites (e.g., Reddit, StackOverflow). Though less accurate than explicitly-collected feedback, it allows feedback models to be trained on much more data. \citet{askell2021general} found that naturally occurring feedback data benefits models larger than 1B parameters, but often has diminishing returns when the number of explicit-collected feedback increases.

\citet{nguyen-etal-2022-make} train a preference model based on rankings on three human-designed objectives: whether the summary has an appropriate topic, length, and quality, combining these three into a single objective using a distance-based ranking loss.
Interestingly, automatic post-editing (APE) systems in MT (\textit{e.g.}, \citet{simard2007rule, correia-martins-2019-simple}), trained on human post-edits with the intent of automatically correcting the output of an MT system, can also be seen as feedback models (albeit non-numerical).

\subsection{Leveraging Feedback Models to Improve Generation}
\label{subsec:Leveraging Preference Models to Improve Generation}

After training a feedback model, we can use it to improve generation almost exactly as we would use human feedback: either by leveraging this feedback model during the training of the generation model, or by incorporating the feedback model during the decoding process.

\subsubsection{Optimizing for Feedback Models}
\label{subsubsec:indirect-optimize}

Similarly to optimizing for human feedback, one possible way to use the feedback model is to optimize model parameters with respect to the feedback it gives. If the feedback model outputs numerical feedback ($\hat{h}_\phi: \mathcal{X} \times \mathcal{Y} \rightarrow \mathbb{R}$) we can define an optimization problem similar to \autoref{eq:optimize-feedback}. However, due to the limitations of feedback models as imperfect proxies, typically a \textit{regularization} term $R$ is introduced to avoid \textit{``overfitting''} to the feedback model \citep{ziegler-etal-2019-fine} (more on this at the end of this section):
\begin{equation}
\label{eq:optimize-feedback-models}
\theta^\star = \argmax_\theta \EX_{x \sim \mathcal{D}} \left [ \hat{h}_\phi(x, M_\theta(x)) - \beta R(\theta) \right ]
\end{equation}
Due to the similarities between both optimization problems, approaches to tackle \autoref{eq:optimize-feedback-models} can be divided into two of the three categories in \S\ref{subsec:direct-decode}: \textbf{joint-feedback modeling} and \textbf{reinforcement learning}.
Recall that while in \S\ref{subsec:direct-decode} we discuss approaches for directly optimizing for human feedback, while this section is focused on cases where a model of human feedback is used instead.

Unlike when using human feedback directly, most works attempt to optimize for feedback models using \textbf{reinforcement learning}.  \citet{gao-etal-2018-april, bohm-etal-2019-better} use the (numerical) feedback collected in other works to train reward and preference models, and use reinforcement learning to optimize against these models, showing that humans preferred their summarization model to other supervised and RL-trained baselines. \citet{ziegler-etal-2019-fine} proposed a similar approach, but trained preference models using feedback collected on the model being improved, and introduced a \textit{KL} regularization term
\begin{equation}
    \label{eq:kl-reg}
    R(\theta) = \log \left [P_\theta(y | x) / P_{\theta_\textrm{SL}}(y | x) \right ]
\end{equation}

to avoid the optimized model deviating too much from the original (supervised) model with parameters $\theta_\textrm{SL}$\footnote{Note that this KL term is different from other algorithm-specific regularization terms, such as the KL terms in PPO~\citep{schulman2017proximal}.}. \citet{stiennon-etal-2020-learning} extended this work, by \textit{scaling} both the summarization and preference models, showing that their model was highly preferred by humans, and generalized better than supervised baselines. \citet{ouyang2022training} also used reinforcement learning with preference models to improve the ability of LLMs to follow instructions, but combined the RL objective with the original pretraining objective to avoid performance regressions in public NLP benchmarks. Other works have also used reinforcement learning with preference models in a similar manner \cite{askell2021general, bai2022training, wu-etal-2021-recursively, nguyen-etal-2022-make}. Underlying all these methods is that generally the model is first trained with imitation-learning on human demonstrations, which improves performance compared to using reinforcement learning directly on the pretrained policy.

\citet{glaese2022improving} compared doing feedback-based imitation learning with human feedback (\S\ref{subsec:direct-optimize}) with doing reinforcement learning with a feedback model, finding that the latter led to a better preference rate and lower rule violation rate.

The \textbf{joint-feedback modeling} with feedback models was explored
by \citet{korbak2023pretraining}, who study pre-training an LLMs with a loss similar to \autoref{eq:conditional-feedback-modeling}, based on feedback from a preference model trained on ranking-based feedback for toxicity. They showed that this leads to models producing less toxic generations, when compared to pretraining a model with vanilla MLE.

In an approach outside these main categories, \citet{peyrard-gurevych-2018-objective} use a scoring function learned from human judgments as a fitness function for a genetic algorithm to generate summaries of input texts.

\subsubsection{Decoding with Feedback Models}
\label{subsubsec:indirect-decode}

As mentioned, feedback models have the advantage that they can be queried cheaply for feedback once trained. Perhaps for this reason, most approaches that leverage feedback models by sampling a large number of candidate generations, and reranking them according to the feedback model:
\begin{align*}
\mathcal{C} = \{\bar{y}_{1}, \cdots&, \bar{y}_{S}\} \;\;\textrm{where}\;\; \bar{y}_i \sim  P_\theta\left (y | x \right) \\
\hat{y} &= \argmax_{\bar{y} \in \mathcal{C}} \hat{h}_\phi(x, \bar{y})
\end{align*}
where $\hat{h}_\phi$ is a trained (numerical) feedback model and $\mathcal{C}$ is a set of $S$ candidate generations given by the model (for example, by sampling from its distribution multiple times).

In machine translation, \citet{fernandes-etal-2022-quality} and \citet{freitag-etal-2022-high} build upon recent advances in automatic quality estimation and evaluation via feedback model training to improve generation. Their framework comprises a candidate generation stage followed by a ranking stage, in which the candidates are scored using quality metrics trained to regress on human assessments (reward models) \citep{rei-etal-2020-comet, rei-etal-2020-unbabels} via $N$-best list reranking or minimum Bayes risk (MBR) decoding \citep{kumar2002minimum}. The highest-scoring candidate is then chosen as the final translation.

\citet{Li_2022} collected a dataset of both numerical and natural language feedback for responses from a QA system, and finetuned a pretrained model to predict both kinds of feedback, using the predicted scores from this feedback model to re-rank the predictions from the model. 

\citet{gao2022scaling} also used this approach to study the scaling properties of feedback models and the problem of "overoptimization" (see below).

Additionally, there are several works combining MT and APE systems at decoding time, in which the output of an MT system is further improved by an APE system \citep{bhattacharyya-etal-2022-findings}.

\paragraph{Feedback Model Overoptimization}

One problem that arises when optimizing a system with a feedback model is that this model is only an imperfect proxy for the ground truth human feedback, therefore, "overoptimizing" for them can lead to systems that receive good feedback from the model, but not humans. This problem is known as the  \textit{overoptimization} problem, and is the main reason for the regularization term in \autoref{eq:optimize-feedback-models}


\citet{gao2022scaling} studies the overoptimization problem in preference models, by both optimizing against it with reinforcement learning (training) and reranking outputs with it (decoding). They found that both using preference models during training or decoding led to similar levels of overoptimization, and that the scale of the generation model helps little with this problem.

\section{Collecting and Using Human Feedback}

\label{sec:collection}


Collecting human feedback can be rather expensive and may present i
ssues for the inexperienced, making it important to leverage existing resources and consider additional data collection carefully. We present an introduction to existing datasets and their collection methods, along with considerations for experimenters creating preference datasets for their own use cases. Additionally, we discuss ethical considerations in the use and collection of human feedback.

In future, richer types of feedback may be collected and we may find ways to make use of this signal. For instance, most existing datasets consist of ranking or numerical scores, but humans prefer to provide richer feedback than labelling \citep{stumpf2007toward, amershi2014power, ghai2021explainable}. Furthermore, variability between human annotators has also not been fully explored \citep{plank2022problem, gehrmann2022repairing}. 

\subsection{Considerations in Data Collection}

\begin{table*}[!ht]
    \centering
    \resizebox{\textwidth}{!}{
    \fontsize{8.5}{9}\selectfont
    \begin{tabular}{@{} l | l | l | l |  l @{}}
        \toprule
        \textbf{Task} & \textbf{Dataset \& their descriptions} & \textbf{Collection method} & \textbf{Platform}  & \textbf{Feedback Type} \\
        \midrule
        Language assistant & \href{https://github.com/anthropics/hh-rlhf}{HH-RLHF}~\citep{bai2022training, perez2022red} &  Explicit & Upwork, MTurk & Ranking \\ 
        \midrule
        Language assistant & \href{https://huggingface.co/datasets/stanfordnlp/SHP}{SHP}~\citep{shp} &  Implicit & Scraped from Reddit & Ranking/Score \\
        \midrule
        Summarization & \href{https://github.com/openai/summarize-from-feedback}{summarize-from-feedback}~\citep{stiennon-etal-2020-learning} &  Explicit & Upwork & Ranking \\
        \midrule
        Question Answering & \href{https://mcgill-nlp.github.io/feedbackqa/}{FeedbackQA}~\citep{Li_2022} & Explicit & MTurk & Score, NL \\
        \midrule 
        \midrule
        Translation & WMT Metrics Shared Task~\cite{freitag-etal-2022-results} & Explicit & Pro translation workflow  & MQM, DA \\
        \midrule
        Summarization & TAC Shared Tasks (\href{https://tac.nist.gov/2008/}{TAC-2008}, \href{https://tac.nist.gov/2009/}{TAC-2009}) & Explicit & \textbf{N/A} & Score \\
        \bottomrule
    \end{tabular}
    }
    \caption{Summary of existing human feedback datasets and their collection methods, which vary along several dimensions. Refer to \autoref{table:feedback} for definitions related to feedback types. A separation is drawn between datasets that were explicitly designed to capture human preferences in a general sense, and datasets designed for more specific use cases, such as MQM/DA datasets in MT. \textbf{N/A} means we could not find information.}
    \label{tab:existing-datasets}
    \vspace{-1.5em}
\end{table*}

 
 There are multiple facets to consider when collecting human feedback data for a generation task; a non-exhaustive list of axes along which data collection can vary is presented below.

\begin{enumerate}[nosep,labelwidth=*,leftmargin=1.2em,align=left,label=\arabic*.]
    \item \textbf{Annotator expertise}: Depending on task and training \citep{snow-etal-2008-cheap, sheng2008, clark-etal-2021-thats, gillick-liu-2010-non, Freitag_2021}, annotators can be domain experts to crowdworkers or even models.
   \item \textbf{Length of engagement}: Involves one-time or long-term collaborations with annotators, with preference datasets often involving extended partnerships \citep{stiennon-etal-2020-learning, bai2022training, Freitag_2021}.
   \item \textbf{Collection method}: Data can be gathered explicitly through experiments or implicitly from online sources/user interactions, with varying noise \citep{kreutzer-etal-2018-neural, Freitag_2021}.
   \item \textbf{Collection platform}: Common platforms include Amazon Mechanical Turk, Upwork, and Scale AI.
   \item \textbf{Annotator demographics}: Different groups may have varying opinions on quality generations; demographics may be collected during data collection.
\end{enumerate}

There is generally a trade-off between the effort needed to create the datasets and the reliability of judgments collected. For higher-stakes applications in specific domains, it may be worth the effort to consult expert annotators in an extended partnership. For general alignment with human preferences, it may instead be prudent to recruit a diverse group of annotators to avoid overfitting to the preferences of specific demographics that may be more accessible in recruitment.

\subsection{Pitfalls and Ethical Considerations of Human Feedback}

Although we have focused on the idealized form of human feedback in \S\ref{sec:background}, actual feedback may be low-quality, contradictory, or adversarial.
\footnote{By adversarial feedback, we mean feedback that intentionally inverts a user's preferences, or is designed to mislead a model in some systematic way, rather than just noisy data.} As discussed in \S\ref{subsec:objective}, we must carefully specify annotation guidelines so that feedback is aligned towards the actual goals for the model \citep{ziegler-etal-2019-fine}. 
Even in the case where human experts are available, different groups of experts may not agree \citep{kahneman-book}.
In this section, we enumerate possible issues with human feedback, most of which are shared with other annotation tasks. We also touch on possible mitigation strategies.

\subsubsection{Subjectivity and variance in judgment}
\label{subsubsec:subjectivity-variance}

Considering $K$ annotators with feedback functions ${h_i}_{i=1}^{K}$, judgments are given on data $\mathcal{D} = {d_1, ..., d_N}$. Inter-rater reliability metrics, such as Cohen's Kappa, Fleiss' Kappa, or Krippendorff's alpha, can assess annotator agreement \citep{krippendorff_a, fleiss_k, cohen_k}. Low reliability may result from unclear tasks or evaluation criteria \citep{gehrmann2022repairing, thomson-reiter-2021-generation}, inherent subjectivity, or multiple plausible interpretations \citep{plank2022problem, nie2020can, Gordon_2022}.

Mitigation strategies include viewing humans as making noisily-rational choices \citep{ghosal2023effect}, learning the reliability level of feedback from multiple humans \citep{yamagata2021reinforcement}, and augmenting evaluation metrics like COMET with confidence intervals \citep{Glushkova_2021, zerva2022disentangling}. Clear annotation guidelines and including rationales with rankings can reduce biases and improve clarity \citep{ziegler-etal-2019-fine}.

\subsubsection{Bias in judgment}

Even if all $K$ annotators agree on a particular judgment for a certain data point, they may all be mistaken. There are well-known biases in human reasoning which may cause all annotators or a large percentage of annotators to be mistaken, or not take evidence into account. Furthermore, even if annotators are technically unbiased in terms of the task they were instructed to evaluate, instructions can be underspecified or lead the annotators to evaluate a slightly different task, leading to the appearance of systematic bias away from the originally intended task \cite{parmar2023dont}.

\textbf{Anchoring/Confirmation bias: } When annotators are presented with a text in isolation, they may fail to consider better alternatives and erroneously label the text as high-quality \citep{bansal2021does}. When asked to generate text, anchoring bias can cause people to write in a different manner than usual \citep{Jakesch2023CoWritingWO, Lehmann2022SuggestionLV}, which may influence what types of suggestions or corrections they give. Mitigation strategies include asking people to rank several diverse outputs and being explicit about the dimensions people are asked to evaluate.

\textbf{Positivity bias: } When giving feedback to learners in traditional RL environments, users tend to give much more positive feedback than negative feedback, which may lead the agent to avoid the goal they are actually trying to reach in these scenarios \citep{Amershi_Cakmak_Knox_Kulesza_2014, knox2013, THOMAZ2008716}.

\subsubsection{Ethical considerations}

Some subjectivity in annotator judgment can arise from differences across cultural or social groups. \citet{santurkar2023opinions} measure opinions in language model generations, demonstrating varying degrees of representation of demographic groups. Several works observe that tuning with human feedback increases the alignment of generated outputs with US liberal views on controversial topics (\citet{perez2022discovering}, \citet{hartmann2023political}). Annotators with different demographic or political backgrounds may disagree on what qualifies as toxic content (\citet{sap2022annotators}, \citet{dingetal2022impact}). This is particularly pronounced when annotators are asked to make ethical judgments, which may vary with cultural context and personal sensibilities (\citet{jiang2022machines}, \citet{talat-etal-2022-machine}).

\citet{steiger2021psychological} survey moderators of toxic content, identifying harms ranging from slight discomfort to lasting psychological harm from the prolonged performance of content moderation tasks; however, the severity and frequency of toxic content examined in content moderation likely exceeds that in other types of human feedback annotation. \citet{shmueli-etal-2021-beyond} identify toxicity classification and generation from open-ended inputs as two NLP annotation tasks that may trigger harmful responses in annotators. They further argue that this moves beyond the ``minimal risk'' requirement for Institutional Review Board exemption in the United States and encourage academic researchers using crowdworker annotation to file for this ethical review of their work.

Media attention has also focused on fair pay for annotators, with one \textit{TIME} article\footnote{\url{https://time.com/6247678/openai-chatgpt-kenya-workers/}} describing annotators paid \$2 USD or less per hour to review toxic content and provide harmfulness annotations for model training.  Research on crowdsourcing (\citet{shmueli-etal-2021-beyond, rothschild2022towards, soratana2022effects, toxtli2021quantifying, hornuf2022hourly}) cautions that inadequate pay, especially for workers in lower-resourced regions, can be a form of worker exploitation.






\section{AI Feedback}
\label{sec:ai-feedback}

Feedback models have been crucial in advancing generation techniques by effectively leveraging feedback. However, they are heavily reliant on human input: for example, \citet{gao2022scaling} found that across various preference model sizes, utilizing fewer than 1,000 comparisons resulted in only minor improvements, with outcomes approximating chance. Moreover, employing static feedback can create consistency and accuracy challenges, as the integration of feedback leads to changes in the model's output distribution. AI-generated feedback, an emerging research area, focuses on harnessing the large language model's own abilities to evaluate and improve its output, enhancing the model without constant human intervention. Two primary approaches have emerged in this domain:

\paragraph{Self AI Feedback} The first approach involves using the same model to provide feedback and improve its output. In this scenario, the model engages in a continuous self-improvement process, learning from its evaluations and refining its capabilities accordingly. Examples of this approach include prompting models to generate harmful responses and revising them for harmlessness~\citep{bai2022constitutional}, or employing rule-based reward models for RLHF fine-tuning~\citep{openai2023gpt4}. Techniques such as iterative output revision through few-shot prompting~\citep{Peng2023CheckYF,shinn2023reflexion,chen2023teaching,paul2023refiner,madaan2023selfrefine,Yang2022Re3GL} have been explored using LLMs like GPT-3.5~\citep{ouyang2022training} and GPT-4~\citep{openai2023gpt4}.
Notably, these techniques demonstrate potential when applied to LLMs trained to adhere to human instructions and align outputs with human preferences. This suggests that incorporating human feedback during training equips AI models to comprehend task requirements better, align outputs with directives, and function as dependable feedback mechanisms, thereby minimizing human intervention. Intriguingly, the capacity to offer valuable AI feedback may depend on the model being trained with human feedback.

\paragraph{External AI Feedback:} The second approach employs a separate model to provide feedback on the model's outputs which is being improved. In this setting, the task model is often paired with a separately trained feedback model~\citep{Yasunaga2020DrRepair,madaan2021graphcorr,Welleck2022SelfCorrect,bai2022constitutional,akyurek2023rl4f}. An advantage of this approach is that the feedback model does not need to be a large, general-purpose model like GPT-4. Thus, training smaller feedback models becomes an attractive alternative when a large amount of feedback is available.

\section{Conclusion}


Recent developments in large language models have emphasised  the need for human feedback to ensure models have desirable behaviour and generate helpful and harmless text.
In this survey paper, we provided an overview of a recent line of research on leveraging (human) feedback to improve natural language generation. 

Despite the relatively infancy of this field, several important observations emerge when comparing all existing works:
\begin{enumerate}
\item Most feedback formats (and available datasets for them) are \textit{underleveraged}: models are mostly optimized using ranking-based or numerical feedback, particularly when using feedback models. However, we have evidence that most forms of feedback could also provide useful signals for improving models, and natural language feedback seems to be a promising format due to its expressiveness.
\item The \textit{``juice’'} in leveraging (human) feedback seems to be in the feedback itself, rather than on the specific method to leverage it. Despite the emphasis given to Reinforcement Learning from Human Feedback (RLHF) by recent popular works, our survey reveals numerous other approaches to leverage feedback, all of which report improvements over non-feedback-augmented baselines, and recent comparative work even suggests that RLHF might be outperformed by simpler, easier to leverage methods \cite{gao2022scaling, rafailov2023direct}. However, a more comprehensive, large-scale study comparing more methods is still lacking.
\item It’s still unclear what role (human) feedback plays in improving the model’s behavior, and how much of it is actually needed: the success of \textit{AI Feedback} seems to hint that we can massively reduce the need for human supervision, and some recent work \cite{zhou2023lima} raises questions if feedback is needed \textit{at all} when a small amount of high-quality data with human instructions is available for supervised learning.
\end{enumerate}

Overall, we hope this survey can help researchers understand the current state of the art, and identify new and existing sources of feedback and ways of leveraging it.

\section*{Acknowledgments}
This work was supported by EU’s Horizon Europe Research and Innovation Actions (UTTER, contract 101070631) and by the projects MAIA and NextGenAI (LISBOA-01-0247-FEDER-045909 and 2022-C05i0102-02).

\bibliography{tacl}
\bibliographystyle{acl_natbib}

\end{document}